%% file: main.tex
\documentclass{article}
\input{lib/packages}
\pdfoutput=1

\begin{document}
    \input{tex/0.front-matter}
    \input{tex/1.introduction}
    \input{tex/2.design-principles}
    \input{tex/3.anomalib}
    \input{tex/4.library-tools}
    \input{tex/5.benchmarks}
    \input{tex/6.conclusion}
    \input{references}
\end{document}

%% file: lib/packages.tex
\usepackage{lib/spconf}

\usepackage{graphicx}
\usepackage{amsmath}
\usepackage{amssymb}
\usepackage{booktabs}
\usepackage{multirow}
\usepackage{algorithm2e}
\RestyleAlgo{ruled} 

\usepackage[frozencache,cachedir=.]{minted}
\usepackage[
    pagebackref,
    colorlinks
]{hyperref}

\usepackage[capitalize]{cleveref}
\crefname{section}{Sec.}{Secs.}
\Crefname{section}{Section}{Sections}
\Crefname{table}{Table}{Tables}
\crefname{table}{Tab.}{Tabs.}

\newcommand{\sota}{state-of-the-art }

\renewcommand{\paragraph}[1]{{\noindent{\textbf{#1}. }}}
\newcommand{\anomalib}{\emph{anomalib }}
\newcommand{\Anomalib}{\emph{Anomalib }}

%% file: tex/0.front-matter.tex
% - - - - - - - - - - - - - - - %
%           Anomalib            %
% - - - - - - - - - - - - - - - %
\title{
    Anomalib: A Deep Learning Library for Anomaly Detection
}

\name{
    Samet Akcay,
    Dick Ameln,
    Ashwin Vaidya,
    Barath Lakshmanan,
    Nilesh Ahuja,
    Utku Genc
}

\address{
    Intel\\
    {
        \tt\small \{
            name.surname
        \}@intel.com
    }
}

\maketitle
\begin{abstract}
    This paper introduces \emph{anomalib}\footnote{\url{https://github.com/openvinotoolkit/anomalib}}, a novel library for unsupervised anomaly detection and localization. With reproducibility and modularity in mind, this open-source library provides algorithms from the literature and a set of tools to design custom anomaly detection algorithms via a plug-and-play approach. \Anomalib comprises state-of-the-art anomaly detection algorithms that achieve top performance on the benchmarks and that can be used off-the-shelf. In addition, the library provides components to design custom algorithms that could be tailored towards specific needs. Additional tools, including experiment trackers, visualizers, and hyper-parameter optimizers, make it simple to design and implement anomaly detection models. The library also supports OpenVINO model-optimization and quantization for real-time deployment. Overall, \emph{anomalib} is an extensive library for the design, implementation, and deployment of unsupervised anomaly detection models from data to the edge.
\end{abstract}


%% file: tex/1.introduction.tex
% - - - - - - - - - - - - - - - %
%       Introduction            %
% - - - - - - - - - - - - - - - %
\section{Introduction}
\label{sec:introduction}
% Anomaly detection is a growing research area in machine learning literature, where the problem domain is characterized by the absence of labeled data during the model's training stage. Instead of leveraging labeled data to learn an implicit mapping to predict a label, the training objective of anomaly detection models is to learn the distribution of the normal data. During inference, examples are compared to this representation of normality to be labeled as normal or anomalous. This unsupervised nature of anomaly detection makes it well suited for real-world applications such as industrial, medical and security, where the type of abnormality is often unknown.

Anomaly detection is a growing research area in machine learning literature, where the goal is to distinguish between normal samples anomalous samples in a dataset. Supervised approaches are generally ineffective for this type of problem due to a lack of sufficient representative samples in the anomalous class. To address this problem, anomaly detection algorithms solely rely on normal samples during the training stage, and identify anomalous samples by comparing against the learned distribution of normal data. This unsupervised nature of anomaly detection makes it well suited for real-world applications such as industrial, medical and security, where a clearly defined anomalous class is often lacking.

% Following the success of deep learning in supervised domains such as image classification and object detection over the past decade, interest in deep-learning based anomaly detection algorithms seen a rise over the past few years. These models, which leverage the expressive power of deep neural network representations to obtain a representation of normality, have pushed the \sota on various public benchmark datasets.

\begin{figure}
    \centering
    \includegraphics[width=.9\linewidth]{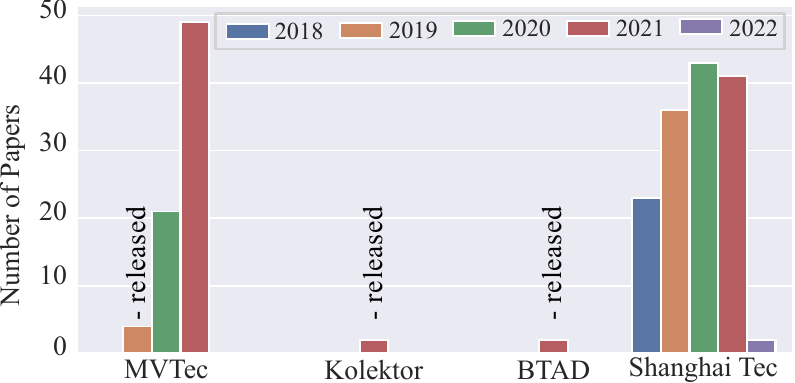}
    \vspace*{-3mm}
    \caption[]{Number of anomaly detection datasets and papers recently released and published in the literature.\footnotemark}
    \label{fig:stats}
\end{figure}
\footnotetext{Paper statistics are taken from \url{https://paperswithcode.com}}

The increasing number of publications and available techniques in the anomaly detection field (Figure \ref{fig:stats}) call for the need for a unified library for benchmarking algorithms. Where supervised tasks have seen various such libraries \cite{mmdetection} \cite{detectron} emerge over the past years, the unsupervised anomaly detection domain lacks similar efforts to date.

%Some authors provide access to the code needed to replicate their experiments, and various third party implementations of popular algorithms can generally be found in public repositories.
Existing anomaly detection libraries focus on single algorithms only, lack performance optimizations, or do not include deep learning techniques \cite{Zhao2019PyOD:Detection}. 
%However, such implementations usually focus on a single algorithm, lack performance optimizations for deployment \cite{Zhao2019PyOD:Detection} or rely on specific dataset preparation steps. 
This makes it challenging to utilize these implementations for out-of-the-box comparison of the most recent algorithms on a given dataset. To address these issues, we introduce \emph{anomalib}, a new library that aims to provide a complete collection of recent deep learning-based anomaly detection techniques and tools.

By collecting different anomaly detection algorithms  and model components in a single library, \emph{anomalib} provides the following advantages: 

\begin{figure*}
    \centering
    \includegraphics[width=\linewidth]{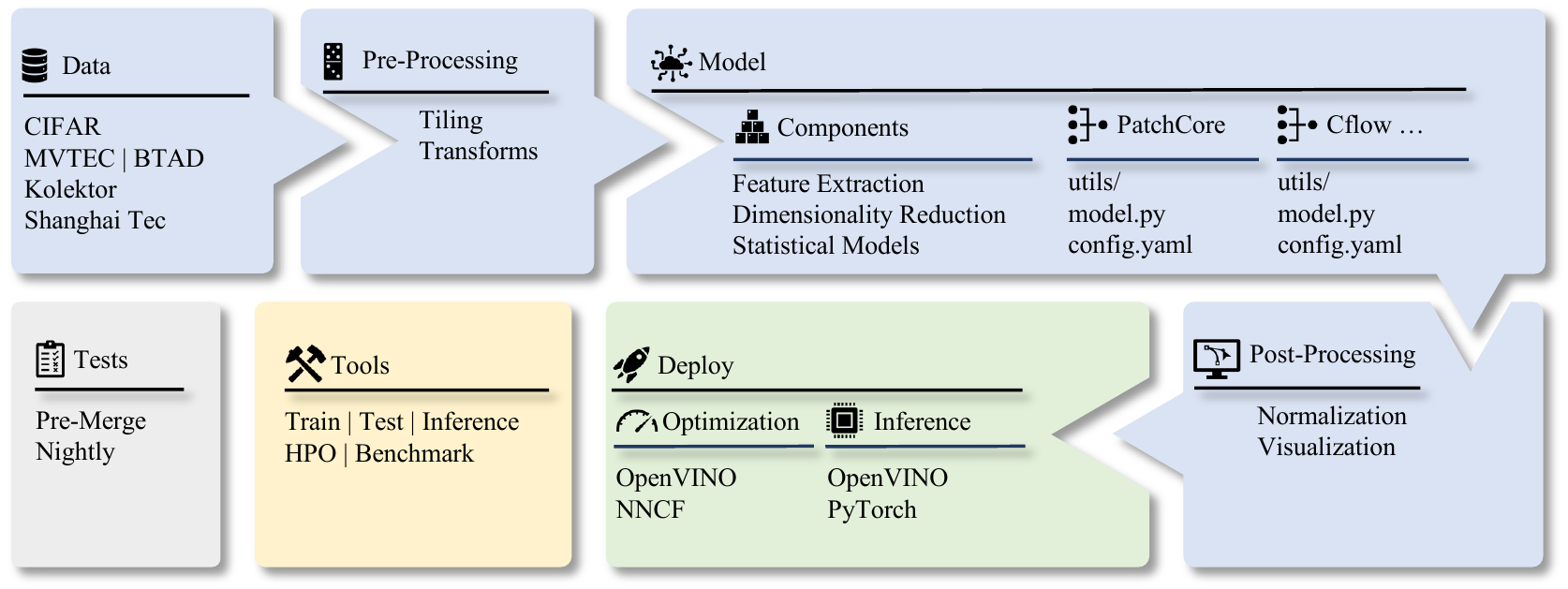}
    \vspace*{-8mm}
    \caption{Architecture of the \Anomalib library from data to deployment.}
    \label{fig:library-components}
\end{figure*}

\begin{itemize}
    \item State-of-the-art anomaly detection models to run benchmark on public and custom datasets.
    \item Modular anomaly model components to design new algorithms via plug-and-play.
    \item CLI-based, configurable entrypoints for training, testing, inference, hyperparameter optimization and benchmarking. 
    \item Inference interfaces, compatible with number of exportable model formats that facilitate real-time local or edge deployment.
\end{itemize}

\noindent Overall, \anomalib is a unified library that provides a set of components and tools that could be used in anomaly detection research and production.

% First of all, collecting the current \sota algorithms in a single library allows for quick benchmarking and valid between-algorithm comparison on custom datasets. Shared dataset adapters and pre-processing methods ensure consistent input data when comparing models. 
% Apart from these basic training and evaluation capabilities, \emph{anomalib} also facilitates deploying trained models locally or on edge devices by exporting trained models to various formats, and by providing inference interfaces that are compatible with the exported model formats. 
% The models implemented in \emph{anomalib} have a modular design, which means that new models can easily be composed by re-using components from other models. This design allows \emph{anomalib} to be used as a research tool for developing and experimenting with new anomaly detection techniques. 
% Finally, \emph{anomalib} provides entrypoints for training, testing, inference, hyperparameter optimization and benchmarking, each of which can be used and configured through a command-line interface. 

% \lipsum[1]

%% file: tex/2.design-principles.tex
% - - - - - - - - - - - - - - - %
%       Design Principles       %
% - - - - - - - - - - - - - - - %
\section{Design Principles}
\label{sec:design}
\Anomalib follows four design principles, each of which is explained below.
%extensibility
%simplicity

\paragraph{Reproducibility}
One of the main goals of the \anomalib library is to compare the performance of different \sota anomaly detection algorithms on public and custom benchmark datasets. To ensure a valid comparison, the algorithm implementations in \emph{anomalib} aim to reproduce the results reported in the original publications. 
% Following the continuous integration approach, the library should be able to yield reproducible results. To validate this, we implemented $100+$ tests run during pre-merge and nightly regression tests. The user can fix the seed of the random number generator to enforce consistent results between different training runs.

\paragraph{Extensibility}
Given the fast pace of progress in the anomaly detection field, it is crucial that new algorithms can be added to the library with minimal effort. To this end, \emph{anomalib} provides several interfaces that developers can implement to make their models compatible with the training and inference entrypoints of the library. 
% All anomaly models are implemented using PyTorch \cite{PyTorch}, such that they could be modified in-place, or used in another code-base. 
% To add extra layer of flexibility, we utilize PyTorch Lightning \cite{Falcon2020PyTorchLightning}, which successfully separates the research/production/boilerplate code and provides number of additional tools for logging, experiment management, and hyper-parameter optimization (HPO).

\paragraph{Modularity}
% Modularity is another aspect of \emph{anomalib} that contributes to the flexibility of the library. 
The library contains several ready-to-use components that can serve as building blocks when creating new algorithms. 
% Examples of such components are feature extraction backbones, dimensionality reduction techniques such as Principal Component Analysis (PCA), and statistical modeling methods such as Kernel Density Estimation (KDE) and K-Nearest Neighbor search (KNN). 
Developers and researchers can re-use these components in a plug-and-play fashion to further reduce implementation efforts and quickly prototype new ideas.
%Another main goal to implement such library is to be able to design new anomaly detection algorithms by utilizing components with plug-and-play approach. The user should be able to stack the components together without concerning the implementational details of a component. 

\paragraph{Real-Time Performance}
% Another goal of \anomalib is to simplify the \emph{data-to-deployment} process. The library therefore allows the user to deploy models in real-time using either GPU or CPU via PyTorch \cite{PyTorch} and OpenVINO \cite{OpenVINO} deployment options, respectively.
A key objective of \anomalib is to reduce the effort of performing inference with trained models. The library therefore provides interfaces to deploy models in real-time using either GPU or CPU via PyTorch \cite{PyTorch} and OpenVINO \cite{OpenVINO} deployment options, respectively.

% Overall, these design principles facilitate \emph{data-to-deployment} approach, where the results of the existing algorithms could be reproduced, new algorithms could be designed and the models could be deployed in real-time.
\noindent By following the design principles outlined above, \anomalib aims to cover the full machine learning model lifecycle from data to deployment, where the results of existing algorithms can be reproduced, new datasets and algorithms can be added, and models can be deployed in real-time.

%% file: tex/3.anomalib.tex
% - - - - - - - - - - - - - - - %
%           \emph{anomalib}            %
% - - - - - - - - - - - - - - - %

% \vspace{-0.25cm}
\section{Anomalib}
\label{sec:components}
This section categorizes the library on a component level,\footnote{For more details, refer to the \href{https://openvinotoolkit.github.io/anomalib/}{documentation}.} each of which is a fundamental step in the \emph{data-to-deployment} workflow shown in Figure \ref{fig:library-components} .

% - - - - - - - - - - - - - - - %
% DATA
\subsection{Data}
The library provides dataset adapters for a growing number of public benchmark datasets both from image and video domains that are widely used in the literature.

\paragraph{Image}
\emph{Anomalib} supports CIFAR-10 \cite{Cifar} for fast prototyping, and MVTec \cite{MVTec}, BTAD \cite{BTAD} and Kolektor \cite{Kolektor} for real-world defect detection applications.
% CIFAR-10 \cite{Cifar} is one of the first datasets used in anomaly benchmarks to initially show a \emph{proof-of-concept} in the experiments. \emph{anomalib} therefore provides this dataset.  The library also has wide range of industrial defect detection datasets. MVTec \cite{MVTec} has become one of the main anomaly detection and localization benchmark thanks to its various defect types and number of available samples. Hence, it is one of the most prevalent dataset used in the library to reproduce and maintain the results. More recently, Mishra \etal have released BTAD \cite{BTAD} dataset, comprising three defect categories. 

\paragraph{Video}
The library supports video datasets such as ShanghaiTec \cite{Luo2017ShanghaiTech}. Currently, video datasets are only supported on a frame-level basis since the existing \anomalib models are optimized for image domain. We plan to support video anomaly detection models in future releases to address this.

\paragraph{Custom}
In addition to the aforementioned public datasets, \anomalib provides a dataset interface for the users to implement custom datasets, on which new and existing \anomalib models can be trained.

\subsection{Pre-Processing}
Pre-processing consists of applying transformations to the input images before training, and optionally dividing the images into (non-)overlapping tiles.

\paragraph{Transforms}
\Anomalib utilizes the \emph{albumentations} \cite{Buslaev2020Albumentations} library for image transformations, which has the advantage of managing the transformation of the ground truth pixel maps together with the input images. In addition to its extensible Python API, \emph{albumentations} allows reading transformation settings from a config file, which is a useful feature for experimentation and HPO.

% \paragraph{Tiling}
% The majority of real-world datasets have high image resolution and are resized before training. Objects with small anomalous regions become even smaller, making detection more challenging and resulting in poor performance. Tiling the input image alleviates the issue since the size of the anomalous regions remains consistent (See Figure \ref{fig:tiling}). 
\begin{figure}
    \centering
    \includegraphics{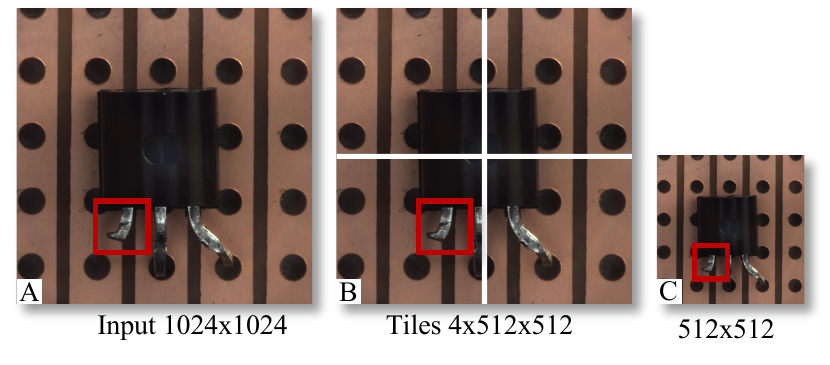}
    \vspace*{-5mm}
    \caption{Tiling vs. resizing the image to the target input size.}
    \label{fig:tiling}
\end{figure}

\paragraph{Tiling}
Due to the high image resolution in many real-world datasets, it is generally required to resize the input images before presenting them to the model. As a side effect, small anomalous regions within the images may lose detail, making it more challenging for the model to detect these regions. Tiling the input images alleviates the issue since the size of the anomalous regions remains consistent (Figure \ref{fig:tiling}).

% \begin{minted}[fontsize=\small]{python}
% >>> from anomalib.pre_processing import Tiler
% >>> tiler = Tiler(tile_size=256, stride=128)
% >>> tiles = tiler.tile(image)
% >>> image.shape, tiles.shape
% Size([3, 512, 512]), Size([9, 3, 256, 256])
% \end{minted}

% \paragraph{Transforms}
% \Anomalib utilizes \emph{albumentations} \cite{Buslaev2020Albumentations} for image transformations since it manages the mask and bounding box transformations  with the image simultaneously. In addition to its extensible Python API, \emph{albumentations} can also read transformations from a config file, which is a useful feature for experimentation and HPO.

% - - - - - - - - - - - - - - - %
% MODEL
\subsection{Model}
% The library comprises growing number of \sota anomaly detection/localization algorithms as well as components to quickly design custom algorithms.
\Anomalib contains a selection of \sota anomaly detection/localization algorithms as well as a set of modular components that serve as building blocks to compose custom algorithms.

\paragraph{Algorithms}
The library is updated periodically with the latest \sota anomaly detection models. Currently available models could be categorized into density estimation \cite{Ahuja2019ProbabilisticDetection, Defard2021PaDiM:Localization, Gudovskiy2021CFLOW-AD:Flows, Roth2021TowardsDetection}, reconstruction \cite{Akcay2018Ganomaly} and knowledge distillation models \cite{Wang2021Student-TeacherDetection}.

\paragraph{Components}
The model components comprise several ready-to-use modules that implement commonly used operations.
% The model components implement several commonly used operations. 
Similar to Scikit Learn \cite{ScikitLearn}, the model components are categorized with respect to their role in anomaly detection models (e.g. feature extraction, dimensionality reduction, statistical modeling). All model components are implemented in PyTorch, which allows running all operations on the GPU and exporting the models to ONNX and OpenVINO. 

Using the model components to implement a custom anomaly detection algorithm is straightforward. For instance, consider an anomaly model that initially extracts features via CNN and performs a dimensionality reduction via Coreset Sampling \cite{Sener2018Coreset}, similar to PatchCore \cite{Roth2021TowardsDetection}. One could import the components as follows:

% We follow a similar classification strategy to Scikit Learn \cite{ScikitLearn} and split the model components into feature extraction, dimensionality reduction, sampling and statistical components. Unlike Scikit Learn, these components are implemented in PyTorch to use GPU during training and export the models to ONNX and OpenVINO for real-time inference, which overall yields significant computational performance gains.

% It is straightforward to use these components to design custom anomaly detection algorithms. For instance, consider an anomaly model that initially extracts features via CNN and performs a dimensionality reduction via Coreset Sampling \cite{Sener2018Coreset}, similar to PatchCore \cite{Roth2021TowardsDetection}. One could import the following components to stack together:
% \vspace{-0.75cm}
\begin{minted}[fontsize=\small]{python}
from anomalib.models.components import (
    FeatureExtractor, KCenterGreedy
)
\end{minted}

% We will keep appending the available components as new algorithm implementations are added to the library.

% \paragraph{Algorithms}
% In addition to model components, \anomalib also contains \sota anomaly detection models, updated periodically. These models could be categorized into density estimation \cite{Ahuja2019ProbabilisticDetection, Defard2021PaDiM:Localization, Gudovskiy2021CFLOW-AD:Flows, Roth2021TowardsDetection}, reconstruction \cite{Akcay2018Ganomaly} and knowledge distillation models \cite{Wang2021Student-TeacherDetection}. 

% For instance, the following line imports PatchCore \cite{Roth2021TowardsDetection} to train a density estimation model.
% \begin{minted}[fontsize=\small]{python}
% from anomalib.models import PatchCore
% \end{minted}

% - - - - - - - - - - - - - - - %
% Post-Processing
\subsection{Post-Processing} 

\begin{figure}
    \centering
    \includegraphics[width=\linewidth]{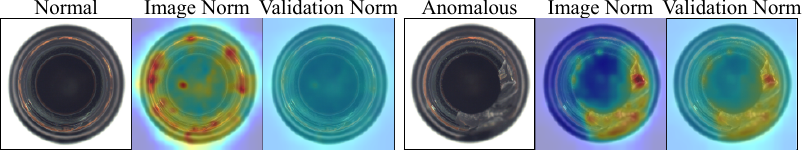}
    \vspace*{-5mm}
    \caption{Normalizing with respect to the validation set results in better visualizations than normalizing with respect to individual images.}
    \label{fig:normalization}
\end{figure}

\paragraph{Normalization}
The range of image-level or pixel-level anomaly scores predicted by the models in \anomalib during inference may vary depending on the model and dataset. To convert the raw anomaly scores into a standardized format, \anomalib normalizes the predicted anomaly scores to the [0,1] range. By default, \anomalib uses min-max normalization with respect to the values observed during validation (Figure \ref{fig:normalization}), but the normalization method can be configured or disabled entirely.

\paragraph{Thresholding}
To help the user choose an anomaly score threshold for their trained models, \anomalib provides an adaptive thresholding mechanism, which optimizes the value of the threshold based on the F1 score during validation. Alternatively, the user can specify a manual threshold. This is useful in cases where insufficient representative validation data is available.

\paragraph{Visualization}
During validation and testing, \anomalib can be configured to show and save visualizations of the predicted anomaly heatmaps and segmentation masks (Figure \ref{fig:visualization}).

% - - - - - - - - - - - - - - - %
% Deploy
\subsection{Deploy}

\paragraph{Optimization}
To achieve faster inference and throughput via quantization and optimization, the library utilizes  OpenVINO \cite{OpenVINO} and Neural Network Compression Framework (NNCF) \cite{nncf}.

\paragraph{Inference}
Trained models can be deployed by relying solely on the library's inference utilities, which allow users to visualize the inference results in a window or save predicted anomaly scores to the file system.
% The future release will also support streaming video. Refer section \ref{sec:inference} for more details.

\begin{table*}[]
\resizebox{\textwidth}{!}{
\begin{tabular}{@{}lcccccccccccccccc@{}}
\toprule
          & Carpet & Grid & Leather & Tile & Wood & Bottle & Cable & Capsule & Hazelnut & Metalnut & Pill & Screw & Toothbrush & Transistor & Zipper & Mean \\ \midrule
% GANomaly  & 0.232  & 0.662 & 0.410  & 0.595& 0.618&  0.393 & 0.496 & 0.313  &  0.497    & 0.277    & 0.521 & 0.409 &  0.350    &  0.348     &  0.389 & 0.434  \\
DFM \cite{Ahuja2019ProbabilisticDetection}               & 0.847  & 0.496 & 0.930  & 0.981 & 0.957 & 0.997 & 0.927 & 0.876  & 0.985    & 0.946    & 0.816 & 0.761 & \textbf{0.964}     & 0.936      &  0.945 &  0.891    \\
STFPM \cite{Wang2021Student-TeacherDetection}     & 0.921  & \textbf{0.983} & 0.921  & 0.961& \textbf{0.994} & 0.998  & 0.946 & 0.717  & 0.996    & 0.984    & 0.492& 0.550 & 0.638      &  0.852     & 0.837  &   0.853   \\
PADIM \cite{Defard2021PaDiM:Localization}             & 0.945 & 0.857 & 0.982   & 0.950 & 0.976 &  0.994  & 0.843 & 0.901 & 0.750   &  0.961   & 0.863& 0.759 & 0.889      &   0.920    &  0.780 & 0.891  \\
CFLOW \cite{Gudovskiy2021CFLOW-AD:Flows}     & \textbf{0.981} & 0.963 &  \textbf{1.000}   &  0.997 & 0.991 & \textbf{1.000}  & 0.969  & 0.970    & \textbf{1.000}      & 0.991      & \textbf{0.981}  & 0.767   &  0.911       &  0.950       &   \textbf{0.986}   &  0.964    \\
PatchCore \cite{Roth2021TowardsDetection}              & 0.979  & 0.962 & \textbf{1.000}   & \textbf{1.000}   & 0.992 &  \textbf{1.000}  & \textbf{0.993} & \textbf{0.976}   & \textbf{1.000}      & \textbf{0.996}    & 0.934&  \textbf{0.947} & 0.947     & \textbf{1.000}        &  0.982 &  \textbf{0.981 } \\ \bottomrule
\end{tabular}
}
\vspace*{-3mm}
\caption{Image Level AUROC Scores for MVTec \cite{MVTec} dataset categories. }
\label{table:image-auc}
\end{table*}

\begin{table*}[]
\resizebox{\textwidth}{!}{
\begin{tabular}{@{}lcccccccccccccccc@{}}
\toprule
          & Carpet & Grid & Leather & Tile & Wood & Bottle & Cable & Capsule & Hazelnut & Metalnut & Pill & Screw & Toothbrush & Transistor & Zipper & Mean \\ \midrule
STFPM \cite{Wang2021Student-TeacherDetection}     & 0.973  & \textbf{0.987} &0.976  &0.964& \textbf{0.958} &  0.968 & 0.940 & 0.957  &  0.979   & 0.965    & 0.867 & 0.938 &  0.148     &   0.759    & \textbf{0.983}  &  0.891    \\
PADIM \cite{Defard2021PaDiM:Localization}     &  0.984 & 0.918 & 0.994 &0.934& 0.947 &  0.983 & 0.965 & 0.984  &   0.978  &  0.970   & 0.957& 0.978 &  \textbf{0.988}     &  0.968     &  0.979 & 0.968     \\
CFLOW \cite{Gudovskiy2021CFLOW-AD:Flows}  & 0.985 & 0.969 & \textbf{0.995} &\textbf{0.966}& 0.926 &  0.984 & 0.965 & \textbf{0.988}  &  \textbf{0.988}   &  0.982   & \textbf{0.987} & 0.982 &  0.985  &   0.930 &  0.980 &  0.974    \\
PatchCore \cite{Roth2021TowardsDetection}  & \textbf{0.988}  & 0.969 & 0.991 &0.961& 0.935 &  \textbf{0.985} & \textbf{0.988} & \textbf{0.988}  &   0.987  &   \textbf{0.989}  &0.981 & \textbf{0.989} &   \textbf{0.988}    &   \textbf{0.982}    & \textbf{0.983}  &  \textbf{0.980}    \\ \bottomrule
\end{tabular}
}
\vspace*{-3mm}
\caption{Pixel Level AUROC Scores for MVTec \cite{MVTec} dataset categories. }
\label{table:pixel-auc}
\end{table*}

% - - - - - - - - - - - - - - - %
% Utilities
\subsection{Utilities}
\Anomalib uses a number of utilities and helper modules to facilitate a complete training and inference pipeline. 

\paragraph{Callbacks}
The library utilizes PyTorch Lightning's callback class \cite{Falcon2020PyTorchLightning} for non-model-specific operations such as normalization, OpenVINO/NNCF compression, timer and visualization to reduce boilerplate code in the model implementations.
% The library fully supports PyTorch Lightning \cite{Falcon2020PyTorchLightning} callbacks and utilizes it's base class to implement custom callbacks such as normalization, OpenVINO/NNCF compression, timer and visualization. 

\paragraph{Metrics}
Commonly reported performance metrics such as AUROC, F1 and PRO are reported after each training run. The metrics are implemented using TorchMetrics \cite{Falcon2020PyTorchLightning}, which allows running the computations on GPU.
% By following the performance metrics used in the literature, the library comprises AUC, F1 and PRO metrics. Since the computation of these metrics are complex, their implementations are done using torchmetrics \cite{Falcon2020PyTorchLightning} to run on GPU.

\paragraph{Logging}
The library supports multiple logging targets such as TensorBoard \cite{Abadi2015TensorFlow:Systems} and wandb \cite{Biewald2020ExperimentBiases} to track experiments. 
The library provides an interface so that the implementation details of the loggers remain abstracted from the user.

% We provide a common base class (\texttt{ImageLoggerBase}) to ensure that the interfaces to log images remain the same across all loggers. This is to ensure that the implementation details for the loggers remain abstracted from the algorithm implementers. 

% The library provides two methods to attach these loggers to the models. The first is by importing the respective logger directly.
% \begin{minted}[breaklines,fontsize=\small]{python}
% from pytorch_lightning import Trainer
% from anomalib.utils.loggers import \emph{anomalib}TensorBoardLogger
% logger = \emph{anomalib}TensorBoardLogger("tb_logs", name="my_model")
% trainer = Trainer(logger=logger)
% \end{minted}

% The second provided approach is to set the \texttt{logger} parameter under \textit{projects} in the \textit{config.yaml} file to \texttt{wandb} or \texttt{tensorboard}. This will automatically initialize the respective logger and pass it to the trainer.

% The implementers can then log the metrics and images from the \texttt{LightningModule}.
% \begin{minted}[fontsize=\small]{python}
% self.log({'loss': loss})
% self.logger.add_image(image, name)
% \end{minted}

%% file: tex/4.library-tools.tex
% - - - - - - - - - - - - - - - %
%       Library Tools           %
% - - - - - - - - - - - - - - - %
\section{Library Tools}
\label{sec:tools}
\subsection{Command Line Interface (CLI)}
\label{ssec:cli}
\Anomalib provides several ready-to-use scripts to train and export models, run inference on trained PyTorch or OpenVINO models, and run Benchmarking and Hyperparameter Optimization experiments. 

\paragraph{Training, Testing and Inference}
\Anomalib provides a set of python scripts for basic training (\texttt{train.py}), testing (\texttt{test.py}) and inference (\texttt{inference.py}) functionality. Each script has a set of command line arguments that can be used to configure the dataset, model and hyperparameter settings. Visualization results and model files will be saved to a file system location specified by the user. The inference entrypoint script supports both PyTorch (\texttt{.ckpt}) and OpenVINO (\texttt{.bin}, \texttt{.xml}) models, depending on the extension of specified model file.

% \paragraph{Training}
% The library has a training script that expects either the name of the algorithm or a configuration file to run the training. The \texttt{model} argument assumes that the requested algorithm is placed in \path{anomalib/models/<name>/model.py}. The \texttt{config} argument is optional and points to the same directory as the \texttt{model.py} if not provided.

% \begin{minted}[breaklines,fontsize=\small]{bash}
% python train.py --model cflow --config config.yaml
% \end{minted}

% \paragraph{Testing}
% The arguments to test are similar to \texttt{train.py} except \texttt{test.py} expects path to a \textit{weight file}.

% \begin{minted}[breaklines,fontsize=\small]{bash}
% python test.py --model stfpm --config config.yaml --weight_file model.ckpt
% \end{minted}

% \paragraph{Inference}
% \label{sec:inference}
% Trained models can be used for inference by running \texttt{inference.py} and providing the file path of the changed model. The extension of the model informs the script to use either PyTorch (.ckpt) or OpenVINO (.onnx, .bin, .xml) backend. The inferencer displays the prediction in a window and/or saves the result to the given path. 
% When using OpenVINO, \textit{meta\_data} file is required to ensure correct thresholding and normalization.

% \begin{minted}[breaklines, fontsize=\small]{bash}
% python inference.py --config config.yaml --weight_file model.[ckpt,onnx,bin,xml] --image_path infer.png --save_path ./results --meta_data config.json
% \end{minted}

\paragraph{Hyperparameter Optimization}
The library provides hyperparameter optimization (HPO) support using a Weights \& Biases (wandb) \cite{Biewald2020ExperimentBiases} plugin. Settings of the HPO sweep such as included parameters, monitored performance metric and number of experiments can be configured from the provided \textit{sweep.yaml} file.% is needed to inform the script regarding the parameters used for optimizing, the metric to optimize for, and the number of experiments to conduct to isolate the best combination of parameters. 
% An example of the sweep configuration is provided in \path{tools/hpo/sweep.yaml}.

% \begin{minted}[breaklines, fontsize=\small]{bash}
% python hpo.py --model padim --model_config config.yaml sweep_config sweep_config.yaml
% \end{minted}

\paragraph{Benchmarking}
% When conducting experiments, it is useful to track the model drift, perform a grid search, and compare model throughput on various devices. 
The library contains a suite of benchmarking scripts for collecting statistical and computational metrics across multiple models or datasets. The entrypoint is a Python script (\texttt{benchmark.py}), which is used to perform a grid search and log the results to TensorBoard, wandb, or a local \texttt{.csv} file. 
% This depends on the \textit{benchmark\_params.yaml}, which holds the configuration for the sweep. 
% In addition, a bash script is provided to collect GPU utilization metrics. 
% \begin{minted}[breaklines, fontsize=\small]{bash}
% sudo -E ./train.sh --config config.yaml
% \end{minted}

% \begin{minted}[breaklines, fontsize=\small]{bash}
% python tools/benchmarking/benchmark.py
% \end{minted}

\subsection{Python API}
\label{ssec:python-api}
In addition to the CLI entrypoints, it is also possible to use the Python API for more flexible use of the library or when designing custom algorithms. The following code block demonstrates how a PatchCore \cite{Roth2021TowardsDetection} model could be trained and test on MVTec bottle \cite{MVTec} category.
% \begin{algorithm}
% \caption{An algorithm with caption}\label{alg:two}
% \vspace{-0.1cm}
\begin{minted}[fontsize=\small]{python}
from anomalib.data import Mvtec
from anomalib.models import Patchcore
from pytorch_lightning import Trainer

datamodule = Mvtec(category="bottle", ...)
model = Patchcore(backbone="resnet18", ...)
trainer = Trainer(...)
trainer.fit(datamodule, model)
trainer.test(datamodule, model)
\end{minted}
% \end{algorithm}

%% file: tex/5.benchmarks.tex
% - - - - - - - - - - - - - - - %
%           Benchmarks          %
% - - - - - - - - - - - - - - - %
\section{Benchmarks}
\label{sec:benchmarks}
\vspace*{-1mm}
Tables \ref{table:image-auc} and \ref{table:pixel-auc} demonstrate image-level and pixel-level AUROC scores benchmarked and averaged on MVTec \cite{MVTec} categories \footnote{Refer to \href{https://github.com/openvinotoolkit/anomalib}{anomalib} for more benchmark results} for the different public models currently implemented in \anomalib. The results were obtained by using \anomalib's benchmarking tool and illustrate how \anomalib can be used to perform a comparative study between different models and dataset categories.

% One of the advantages of \anomalib is that benchmarking results for all models can be replicated by anyone with access to the \anomalib code.

% Table \ref{table:throughput} shows a comparison of PyTorch and OpenVINO inference throughputs run on NVIDIA GeForce RTX 3090 GPU and Intel Core i9-10980XE CPU, respectively.

\begin{figure}
    \centering
    \includegraphics[width=\linewidth]{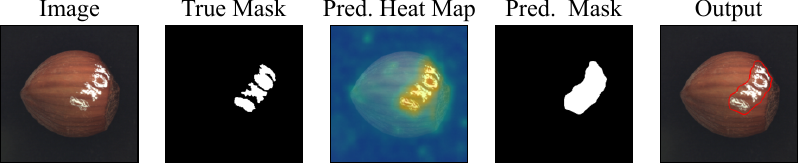}
    \vspace*{-7mm}
    \caption{An exemplary output produced by the \anomalib visualization tools.}
    \label{fig:visualization}
\end{figure}

% \begin{table}[H]
% \resizebox{\linewidth}{!}{
% \begin{tabular}{cccc}
% \hline
%       & PyTorch / GPU (FPS) & OpenVINO / CPU (FPS) & Speedup \\ \hline
% STFPM \cite{Wang2021Student-TeacherDetection} & 6.369         & 12.038         & 89\%    \\
% PADIM \cite{Defard2021PaDiM:Localization} & 15.103        & 20.720         & 37\%    \\ \hline
% \end{tabular}
% }
% \caption{Comparison of PyTorch and OpenVINO throughputs. Averaged across all MVTec Categories.}
% \label{table:throughput}
% \end{table}

%% file: tex/6.conclusion.tex
% - - - - - - - - - - - - - - - %
%           Conclusions          %
% - - - - - - - - - - - - - - - %
\section{Conclusions}
We introduce \emph{anomalib}, a comprehensive library for training, benchmarking, deploying and developing deep-learning based anomaly detection models. The library provides a set of tools that allow quick and reproducible comparison of different anomaly detection models on any dataset, as illustrated by the benchmarking experiments described in this paper. We release it as an open-source package with the aim of constantly updating with the latest \sota techniques in the field, and welcome the community contribution. In future work we plan to extend \anomalib to other domains such as audio, video, and 3-dimensional data.

\label{sec:conclusions}

%% file: references.tex
%%%%%%%%% REFERENCES
% \vfill\newpage
\clearpage
{
    \small
    \bibliographystyle{ref/IEEEbib}
    \bibliography{references}
}